\RequirePackage{xcolor}
\documentclass{bmvc2k}

\usepackage{color}

\title{Robust Adversarial Perturbation on Deep Proposal-based Models}

\addauthor{Yuezun Li}{yli52@albany.edu}{1}
\addauthor{Daniel Tian}{dtian19@berkshireschool.org}{2}
\addauthor{Ming-Ching Chang}{mchang2@albany.edu}{1}
\addauthor{Xiao Bian}{xiao.bian@ge.com}{3}
\addauthor{Siwei Lyu}{slyu@albany.edu}{1}

\addinstitution{
 University at Albany, \\ State University of New York, USA
}
\addinstitution{
 Berkshire School, \\Massachusetts, USA
}
\addinstitution{
 GE Global Research Center, \\Niskayuna, New York, USA
}

\runninghead{Li, et al}{R-AP on Deep Proposal-based Models}

\def\eg{\emph{e.g}\bmvaOneDot}

\def\etal{\emph{et al}\bmvaOneDot}
\def\ie{\emph{i.e}\bmvaOneDot}
\usepackage{color}

\usepackage{algorithm}
\usepackage[noend]{algpseudocode}
\usepackage[colorinlistoftodos]{todonotes}
\usepackage{wrapfig}

\begin{document}

\maketitle
\begin{abstract}  
Adversarial noises are useful tools to probe the weakness of deep learning based computer vision algorithms. In this paper, we describe a robust adversarial perturbation
(R-AP) method to attack deep proposal-based object detectors and instance segmentation algorithms. Our method focuses on attacking the common component in these algorithms, namely Region Proposal Network (RPN), to universally degrade their performance in a black-box fashion. To do so, we design a loss function that combines a label loss and a novel shape loss, and optimize it with respect to image using a gradient based iterative algorithm. 
Evaluations are performed on the MS COCO 2014 dataset for the adversarial attacking of 6 state-of-the-art object detectors and 2 instance segmentation algorithms. 
Experimental results demonstrate the efficacy of the proposed method. 
\end{abstract}

\section{Introduction}
\label{sec:intro}

Deep learning based algorithms achieve superior performance in many problems in computer vision, including image classification, object detection and segmentation. However, it has been recently shown that algorithms based on Convolutional Neural Network (CNN) are vulnerable to {\em adversarial perturbations}, which are  intentionally crafted noises that are imperceptible to human observer, but can lead to large errors in the deep network models when added to images.  To date, most existing adversarial perturbations are designed to attack CNN image classifiers, \eg, \cite{zeng2017adversarial,szegedy2013intriguing,goodfellow2014explaining,kurakin2016adversarial,papernot2016limitations,moosavi2016deepfool,evtimov2017robust,moosavi2017universal}. 

Recently, attention has been shifted to finding effective adversarial perturbation to CNN-based object detectors \cite{xie2017adversarial,lu2017adversarial}. Compared to image classification, finding effective perturbations for object detectors is more challenging, as the perturbation should affect not just the class label, but also the location and size of each object within the image. 
Existing methods \cite{xie2017adversarial,lu2017adversarial} mostly design specific loss functions based on the final prediction to disturb object class labels. As such, these methods are model dependent, which require detailed knowledge of the network architectures.

\begin{figure*}[t]
	\centering
	\includegraphics[width=0.95\linewidth]{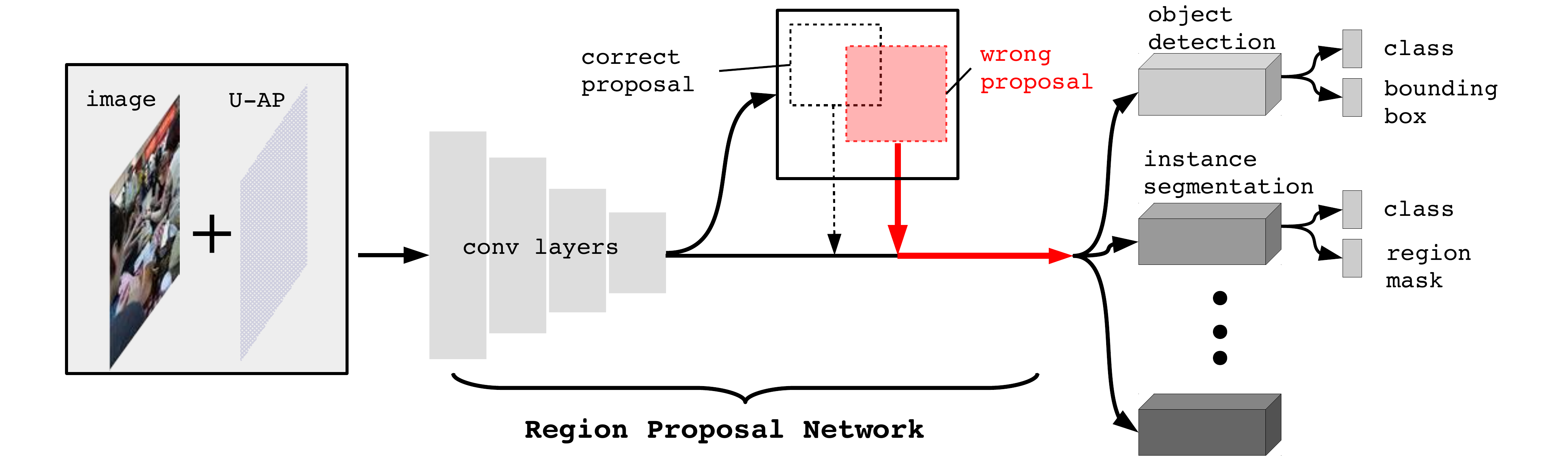}
	\vspace{-0.2cm}
	\caption{\small \em Overview of the Robust Adversarial Perturbation (R-AP) method. Our method attacks Region Proposal Network (RPN) \citep{faster-rcnn} in deep proposal-based object detectors and instance segmentation algorithms.}
	\label{fig:overview}
\end{figure*}

In this work, we develop a {\em Robust Adversarial Perturbation} (R-AP) method to universally attack deep proposal-based models that are fundamental to majority of object detectors and instance segmentation algorithms. 
Our method is based on the fact that a majority of recent object detectors and instance segmentation algorithms, \eg, \cite{rfcn,faster-rcnn,FCIS,mask-rcnn} use a Region Proposal Network (RPN) \citep{faster-rcnn} to extract object-like regions, known as {\em proposals}, from an image and then process the proposals further to obtain object class labels and bounding boxes in object detection, and the
instance class labels and region masks in instance segmentation.
If a RPN is successfully attacked by an adversarial perturbation, such that no correct proposals generated, the subsequent process in the object detection or instance segmentation pipeline will be affected. Figure \ref{fig:overview} overviews the proposed R-AP method. 
The investigation of adversarial perturbation on deep proposal-based models can lead to further understanding of the vulnerabilities of these widely applied methods. The efforts can also aid improving the reliability and safety of the derived technologies, including  computer vision guided autonomous cars and visual analytics.

The proposed R-AP method attacks a RPN based on the optimization of two loss functions: {\em (i)} the {\em label loss} and {\em (ii)} the {\em shape loss}, each of which targets a specific aspect of RPN. 
First, inspired by recent methods \cite{xie2017adversarial,lu2017adversarial} that attacks CNN-based object detectors, we design the label loss to disturb the label prediction (which indicates whether a proposal represents an object or not). 
Second, we also design a shape loss, which attacks the shape regression step in RPN, so that even if an object is correctly identified, the bounding box cannot be accurately refined to the object
 shape. 
 Note that our R-AP method can be combined with existing adversarial perturbation method such as \cite{xie2017adversarial} to jointly attack corresponding network, since R-AP specifically focuses on attacking RPN, which is the intermediate stage of network compared to others which target the entire network. 

Experimental validations are performed on the MS COCO 2014 \cite{Lin2014MicrosoftCC}, the current largest dataset used for training and evaluating mainstream object detectors and instance segmentation algorithms. Our experimental results demonstrate that the proposed R-AP attack can significantly degrade the performance of several state-of-the-art object detectors and instance segmentation algorithms, with minimal perceptible artifacts to human observers. 

Our contributions are summarized in the following:
\begin{itemize}
	\item To the best of our knowledge, this is the first work to thoroughly investigate the effects of adversarial perturbation on RPN, which universally affects the performance of deep proposal-based object detectors and instance segmentation algorithms. 
	\item In contrast to previous attack paradigms that only disturb object class label prediction, our method not only disturbs the proposal label prediction in RPN, but also distracts the shape regression, which can explicitly degrade the bounding box prediction in proposal generation.
\end{itemize}

\section{Related Work} 

{\flushleft \bf Deep proposal-based models} follow a common paradigm of two steps --- proposal generation and proposal refinement. A majority of recent object detectors and instance segmentation algorithms are deep proposal-based models.
For object detection \cite{faster-rcnn,rfcn}, a Region Proposal Network (RPN) generates object proposals, which are refined in subsequent network modules for the exact bounding boxes and class labels. 
The state-of-the-art instance segmentation \cite{mask-rcnn,FCIS} can be viewed as an extended version of object detection, which also use RPN to generate object proposals and refine them to semantic mask of objects.

{\flushleft \bf Region Proposal Network (RPN)} is a CNN-based model for object proposal generation. A RPN starts with a (manually specified) fixed size of multi-scale anchor boxes for each cell in feature map. At training phase, each anchor box is matched to ground truth. If the overlap between an anchor box and a ground truth is greater than threshold, this anchor box will be marked as positive example, otherwise negative example. Moreover, the shape offset between positive anchor box and the matched ground truth is recored for bounding box shape regression. At testing phase, the label and offset predictions of all anchor boxes are generated within a single forward. Compared to the selective search method \citep{selective-search} used in RCNN \citep{rcnn}, RPN is much more efficient and accurate, such that it is widely used in current deep models to provide proposals. 

{\flushleft \bf Adversarial perturbation} is an intentionally crafted noise that aims to perturb deep learning based models with minimal perception distortion to the image. Many methods \cite{evtimov2017robust,szegedy2013intriguing,goodfellow2014explaining,kurakin2016adversarial,papernot2016limitations,moosavi2016deepfool,moosavi2017universal,zeng2017adversarial} have been proposed to fool image classifiers. Szegedy \etal \cite{szegedy2013intriguing} first described this intriguing property and formulated adversarial perturbation generation as an optimization problem. Goodfellow \etal \cite{goodfellow2014explaining}
proposed an optimal max-norm constrained perturbations, referred as ``fast gradient sign method'' (FGSM), to improve the running efficiency. Kurakin \etal \cite{kurakin2016adversarial} proposed a ``basic iterative method'' which generates perturbation iteratively using FGSM. Papernot \etal \cite{papernot2016limitations} constructed an adversarial salieny map to indicate the desired places that can be affected efficiently. The DeepFool of Moosavi \etal \cite{moosavi2016deepfool} further improves the effectiveness of adversarial perturbation. Moosavi \etal \cite{moosavi2017universal} discovered the existence of image agnostic adversarial perturbations for image classifier. 

Recently, adversary attack on object detectors has attracted many attentions. 
Lu \etal \cite{lu2017adversarial} attempted to generate adversarial perturbations on ``stop'' sign and ``face'' to mislead corresponding detectors. Xie \etal \cite{xie2017adversarial} proposed a dense adversarial generation method to iteratively incorrect predictions of object detectors. However, these methods are task-specific, which designs loss functions based on
 the final predictions. They do not address the adversarial perturbation universally. In contrast, we focus on attacking RPN, a common component of deep proposal-based models, to universally degrade their performance without knowing the details of their architecture.

\begin{figure*}[t]
	\centering
	\includegraphics[width=\linewidth]{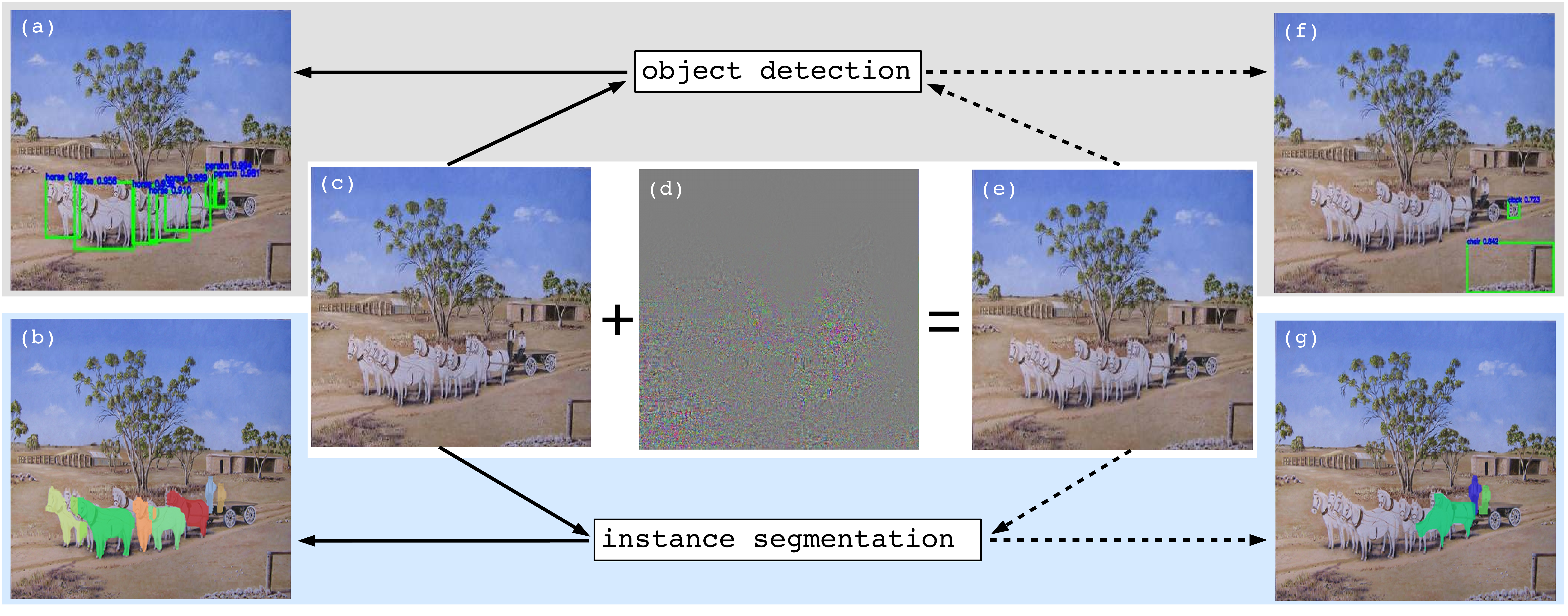}
	\caption{\small \em Illustration of performing R-AP on the object detector Faster-RCNN \cite{faster-rcnn} and the  instance segmentation algorithm FCIS \cite{FCIS}. (c) is the original image. (d) is the noise generated from R-AP, amplified by a factor of $10$ for better visibility. (e) is the perturbed image by adding (c) and (d). 
		(a,f) are the Faster-RCNN object detection of (c,e), and (b,g) are the FCIS instance segmentation of (c,e), respectively.}
	\label{fig:black_demo}
\end{figure*}
\section{Method}

The proposed method attacks deep proposal-based object detectors and instance segmentation algorithms by adding minimal adversarial noises to the input image that can effectively disturb the predictions of Region Proposal Network (RPN). Given an input image and a pre-trained RPN, we design a specific objective function, a combination of two terms --- the label loss and the shape loss, to calculate the adversarial perturbation. In particular, we optimize this objective function
 with respect to image using an iterative gradient based method.

Note all mainstream deep proposal-based object detectors and instance segmentation algorithms rely on a few standard RPNs to provide proposals for subsequent processes. Once the RPN is disturbed, the performance of these deep models is naturally degraded. As such, our R-AP method is suitable in nature for black-box attack to these models, \ie, without the need to know their implementation details. 
Inspired by \cite{xie2017adversarial}, we generate adversarial perturbations for different RPN and 
combine them together to improve the robustness of black-box attack. Figure \ref{fig:black_demo} illustrates an example of R-AP attack on object detection
 and instance segmentation.

Section \ref{subsec:Formulation} describes the notations and the general paradigm of label loss for generating adversarial perturbations. Then we introduce our new shape loss that can explicitly disturb proposal shape regression. Section \ref{subsec:generation} presents the details of
 iterative adversarial perturbation generation scheme.

\subsection{Notations and Problem Formulation}
\label{subsec:Formulation}

Denote
 $\cal I$ as the input
  image that contains $n$ ground truth bounding boxes for objects $\{ \bar{b}_i = (\bar{x}_i,\bar{y}_i,\bar{w}_i,\bar{h}_i) \}_{i =1}^n$, where $\bar{x}_i,\bar{y}_i,\bar{w}_i,\bar{h}_i$ are the $x$- and $y$-coordinate of the box center point, the width and height of bounding box $\bar{b}_i$, respectively.
Let ${\cal F}_\theta$ denote a Region Proposal Network (RPN) with model parameters $\theta$. Let
 ${\cal F}_\theta({\cal I}) = \{(s_j, b_j)\}_{j =1}^m$ denotes the set of $m$ generated proposals with input image $\cal I$, where $s_j$ denotes the confidence score (probability after sigmoid function) of $j$-th proposal and $b_j$ is the bounding box of $j$-th proposal. Let $b_j = ({x}_j,{y}_j,{w}_j,{h}_j)$, where ${x}_j,{y}_j,{w}_j,{h}_j$ are the $x$- and $y$-coordinate of the box 
 center point, the width and height of bounding box $b_j$, respectively. 

Our goal is to seek an minimal adversarial perturbation added to image $\cal I$ to fail a RPN
. The adversarial perturbation generation can be casted as an optimization problem of specific designed loss. In our method, we design the loss as the summation of {\em (i)} 
the {\em label loss} $L_{label}$, which is a general paradigm used in previous methods to disturb label prediction, and {\em (ii)} the {\em shape loss}
 $L_{shape}$, which is our newly proposed term 
 to explicitly disturb bounding box shape regression.
As {\em Peak Signal-to-Noise Ratio} (PSNR) is an approximation of human perception of image quality, we employ it to evaluate the distortion of adversarial perturbation. Less perturbation results in higher PSNR.
Throughout this work, we assume the model parameters $\theta$ of PRN are fixed, and the R-AP algorithm generates a perturbed image $\cal I$ by optimizing the following loss as 
\begin{equation}
\begin{array}{ll}
\min_{\cal I} \;\;\; L_{label}({\cal I}; {\cal F}_\theta) + L_{shape}({\cal I}; {\cal F}_\theta), \;\; \textrm{s.t.} \;\; \textrm{PSNR}({\cal I}) \ge \epsilon,
\end{array}
\label{equ:total}
\end{equation}
where $\textrm{PSNR}({\cal I})$ denotes the PSNR of luminance channel in image ${\cal I}$, $\epsilon$ is the lower bound of PSNR.
We describe the label loss $L_{label}$ and shape loss $L_{shape}$ in sequel.


\noindent{\bf Label Loss}.
The label loss $L_{label}$ 
is designed to disrupt the label prediction of proposals, which is in analogy to existing 
adversarial perturbation methods \citep{lu2017adversarial,xie2017adversarial} for object detectors. Denote $z_j \in \{0,1\}$ as the indicator of $j$-th proposal, where $z_j = 1$ means that $j$-th proposal is positive, otherwise negative. We let $z_j = 1$ if (1) the bounding box intersect-over-union (IoU) of $j$-th proposal with an arbitrary ground truth object is greater than a preset threshold $\mu_1$; (2) the confidence score of $j$-th proposal is greater than another preset threshold $\mu_2$, otherwise we set $z_j = 0$. The above rule can be formulated 
as $z_j = 
1$, if $\exists \; i$, $\textrm{IoU}(\bar{b}_i, b_j) > \mu_1$ and $s_j > \mu_2$, and 0 otherwise.

Note that RPN initially generates a large amount of proposals, and in R-AP, we only focus on disturbing positive proposals as they are the key to the subsequent algorithms. The label loss $L_{label}$ is given by
\begin{equation}
\begin{array}{ll}
L_{label}({\cal I}; {\cal F}_\theta) = \sum_{j = 1}^m z_j \log (s_j).
\end{array}
\label{equ:label-loss}
\end{equation}
In other words, minimizing this loss is equivalent to decreasing confidence score of positive proposals.


\begin{algorithm}[H]
        \caption{\small \em Adversarial Perturbation Generation}
	\label{alg:U-AP}
	\small{
		\begin{algorithmic}[1]
			\Require {RPN model ${\cal F}_\theta$; input image $\cal I$; maximal iteration number $T$.}
			\State ${\cal I}_0 = {\cal I}, t = 0$;
			\While{$t < T$ and $\sum^m_{j=1} z_j \neq 0$}
			\State $\hat{p}_t = \nabla_{{\cal I}_t}(L_{label} + L_{shape}) $;
			\State $p_t = \frac{\lambda}{||\hat{p}_t||_2} \cdot \hat{p}_t$; \Comment{$\lambda$ {\em is a fixed scale parameter}}
            \State ${\cal I}_{t+1} = \textrm{clip}({\cal I}_t - p_t)$;
            \If {PSNR$({\cal I}_t) < \varepsilon$}
            \State break
            \EndIf
            \State $t = t + 1$;
			\EndWhile
            \State $p = {\cal I}_t - {\cal I}_0$;
			\Ensure adversarial perturbation $p$
	\end{algorithmic}}
\end{algorithm}

\smallskip
\noindent{\bf Shape Loss}. Shape regression is an important step to refine the bounding box of object detections or proposals. Specifically, in RPN, shape regression is used to adjust the anchor boxes to the ground truth bounding boxes of the object by minimizing the offset between them. Therefore, we design a specific shape loss to explicitly disturb the bounding box shape regression. 
Let $\Delta x_j, \Delta y_j, \Delta w_j, \Delta h_j$ be the predicted $x$- and $y$- coordinate center location offsets, width and height offset of bounding box $b_j$, respectively.
To explicitly disturb the
 shape regression, we define a new loss function $L_{shape}$ as
\begin{equation}
\begin{array}{ll}
L_{shape}({\cal I}; {\cal F}_\theta) = \sum_{j =1}^m z_j ((\Delta x_j - \tau_{x})^2+(\Delta y_j - \tau_{y})^2+(\Delta w_j - \tau_{w})^2 + (\Delta h_j - \tau_{h})^2),
\end{array}
\label{equ:offset-disturb}
\end{equation}
where $\tau_{x}, \tau_{y}, \tau_{w}, \tau_{h}$ are large offsets defined to substitute
 the real offset between anchor boxes and matched ground truth bounding boxes. We are only concerned about the predicted offset of positive proposals, as it is inappropriate to consider the bounding boxes of negative proposals
 . By minimizing Eq. \eqref{equ:offset-disturb}, the R-AP method
  forces predicted offset $\Delta x_j, \Delta y_j, \Delta w_j, \Delta h_j$ approaching $\tau_{x}, \tau_{y}, \tau_{w}, \tau_{h}$ respectively, such that the shape of bounding box $b_j$ will be incorrect.


\subsection{The Robust Adversarial Perturbation (R-AP) Algorithm}
\label{subsec:generation}

To generate the proposed 
R-AP, we optimize Eq. \eqref{equ:total} using an iterative gradient descent scheme, as mentioned in \cite{xie2017adversarial}. Let $t$ denote iteration number. We calculate the gradient of $L_{label} + L_{shape}$ with respect to image $\cal I$ at $t$ as $\hat{p}_t$.
We normalize $p_t = \frac{\lambda}{\|\hat{p}_t\|_2} \cdot \hat{p}_t$ to keep perturbation minimal perceptive and stability of each iteration,
where $\lambda$ is a fixed scale parameter, $\|\cdot\|_2$ is $L2$ norm metric. Then image ${\cal I}_{t+1}$ is updated by ${\cal I}_t - p_t$ and we clip the pixel value back to $[0, 255]$ at the end of
 each iteration. The process
  is repeated until
   (1) the maximum iteration number $T$ is reached, or (2) positive proposals cease to exist, \ie, $\sum^m_{j=1} z_j = 0$, or (3) {\em Peak Signal-to-Noise Ratio} (PSNR) is less than a threshold $\varepsilon$. The algorithm of 
   adversarial perturbation generation is listed in Algorithm \ref{alg:U-AP}.

Note that R-AP is not mutually exclusive to other adversarial perturbation method such as \cite{xie2017adversarial} for object detectors. For instance, we can combine R-AP with method in \cite{xie2017adversarial} to generate more effective adversarial perturbations, since our loss function is based on a different stage of networks compared to other state-of-the-arts algorithms.

\begin{table}[t]
	\small
	\centering
    \caption{\small \em Performance of R-AP on 6 state-of-the-art object detectors at mAP $0.5$ and $0.7$. Lower value denotes better attacking performance.}
    \vspace{0.2cm}
	\begin{tabular}{|l | l| l| l| l| l ||l|}
    \hline
		& {\bf FR-v16} & {\bf FR-mn} & {\bf FR-rn50} & {\bf FR-rn101} & {\bf FR-rn152} & {\bf RFCN} \cite{rfcn} \\
		\hline
		{\bf origin} & 59.2/47.3 & 47.1/32.6 & 59.5/49.4 & 63.5/53.6 & 64.8/54.5 & 60.1/50.0\\
		\hline
		{\bf random} & 58.7/46.5 & 46.5/32.6 & 59.6/48.9 & 63.2/53.2 & 64.6/54.4 & 59.9/49.6\\
		\hline
		{\bf v16} ($p_1$) & \bf 5.1/3.1 & 34.8/22.2 & 47.9/36.8 & 52.7/42.4 & 55.5/45.0 & 54.5/43.8\\
		\hline
		{\bf mn} ($p_2$) & 56.8/44.4 & \bf 11.0/6.1 & 56.7/45.2 & 60.6/50.2 & 62.3/51.4 & 57.5/46.6\\
		\hline
		{\bf rn50} ($p_3$) & 53.8/41.2 & 39.5/25.7 & \bf 10.5/6.6 & 52.8/42.2 & 55.9/44.7 & 53.7/42.6\\
		\hline
		{\bf rn101} ($p_4$) & 54.8/42.6 & 41.0/27.4 & 50.0/39.2 & \bf 16.8/11.0 & 56.0/45.3 & 52.0/40.4\\
		\hline
		{\bf rn152} ($p_5$) & 55.0/41.9 & 41.8/27.4 & 49.8/38.3 & 53.6/42.2 & \bf 17.3/10.6 & 54.4/42.9\\
		\hline
		${\bf P} = \alpha \cdot \sum^5_{i=1} p_i $ & 37.5/25.6 & 26.4/16.5 & 31.3/21.3 & 37.9/27.2 & 41.4/30.1 & \bf 47.0/35.9\\
		\hline
	\end{tabular}
    \label{table:detection-perform}
\end{table}

\section{Experimental Results}
\label{sec:experiment}

In this section, we report the experimental evaluation of R-AP on several state-of-the-art object detectors and instance segmentation algorithms. Section \ref{subsec:dataset} describes the dataset and evaluation metric. Section \ref{subsection:settings} describes the R-AP settings in all experiments. Section \ref{subsection:obj_det} and section \ref{subsection:inst_seg}  presents the R-AP attack experiments on object detectors and instance segmentation algorithms respectively.

\subsection{Dataset}
\label{subsec:dataset}

The performance of the R-AP is evaluated on MS COCO 2014 dataset \cite{Lin2014MicrosoftCC}, which is a large scale dataset containing 80 object categories for multiple tasks, including object detection and instance segmentation.
Experiments are conducted on a subset (random 3000 images) of the MS COCO 2014 validation set and evaluated using ``mean average precision'' (mAP) metric \citep{everingham2010pascal} at intersection-over-union (IoU) threshold $0.5$ and $0.7$.

\subsection{R-AP Settings}
\label{subsection:settings}

We generate adversarial perturbations for five different RPN architectures: vgg16 ({\bf v16}) \cite{simonyan2014very}, mobilenet ({\bf mn}) \cite{howard2017mobilenets}, resnet50 ({\bf rn50}), resnet101 ({\bf rn101}) and resnet152 ({\bf rn152}) \cite{he2016deep}, where the adversarial perturbations are denoted $p_1$, $p_2$, $p_3$, $p_4$, $p_5$, respectively. These RPN architectures are extracted from Faster-RCNN ({\bf FR}) object detectors \cite{faster-rcnn} implemented by \cite{chen17implementation}. We also generate Gaussian noise ({\bf random}) as a perturbation baseline in comparison to demonstrate the effectiveness of R-AP.
Inspired by \cite{xie2017adversarial}, we accumulate these perturbations as ${\bf P} = \alpha \cdot \sum^5_{i=1} p_i$, where $\alpha$ is a scale parameter to control the distortion.

The following parameter values are used throughout the paper: overlap threshold $\mu_1 = 0.1$, confidence score threshold $\mu_2 = 0.4$, offset $\tau_{x} = \tau_{y} = \tau_{w} = \tau_{h} = 10^5$, scale parameters $\lambda = 30, \alpha=0.5$, and maximum iteration number $T=210$. In general, typical values of PSNR in lossy image compression is between $30$ and $50$ dB \cite{telagarapu2011image}. Therefore, we set $\varepsilon = 30$ dB as the lower bound of PSNR. In our experiments, all perturbations are terminated at maximum iteration number $T$ with PSNR $>\varepsilon$.

\begin{figure*}[t]
	\centering
\includegraphics[width=0.95\linewidth]{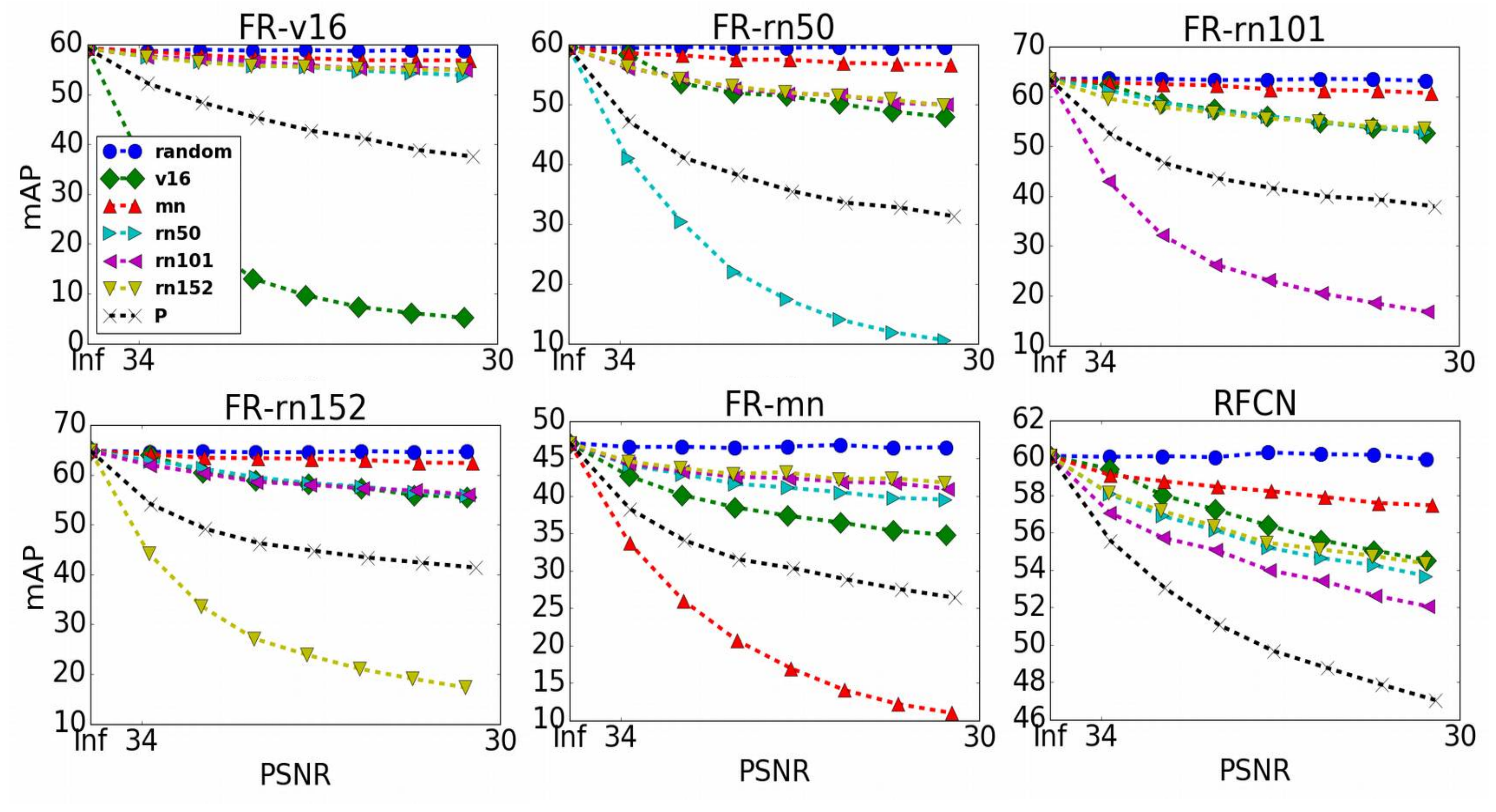}
	\caption{\small \em Illustration of R-AP performance under different PSNR value at mAP 0.5 on 6 object detectors.}
	\label{fig:det_plot_dB}
\end{figure*}

\subsection{Object Detection}
\label{subsection:obj_det}
We study six state-of-the-art object detectors, including five Faster-RCNN based methods {\bf FR-v16}, {\bf FR-mn}, {\bf FR-rn50}, {\bf FR-rn101}, {\bf FR-rn152} with various base networks, and the Region Fully Convolutional Network {\bf RFCN} \cite{rfcn}. We denote FR-v16 as vgg16 based Faster-RCNN in short, and the others are named in the same way accordingly throughout the paper. 

Table \ref{table:detection-perform} illustrates the performance of R-AP generated from different RPN on the six object detectors, where we report mAP metric at 0.5 and 0.7. The $2^{nd}$ row of Table \ref{table:detection-perform} (random) shows the added Gaussian noise as perturbation for a baseline comparison, and the result shows that the performance degradation is $<1\%$. In contrast, the R-AP generated from {\em v16, mn, rn50, rn101, rn152} can each reduce the performance of the object detectors by a larger amount. Since we extract {\em v16, mn, rn50, rn101, rn152} based RPN from {\em FR-v16, FR-mn, FR-rn50, FR-rn101, FR-rn152} detectors respectively, the R-AP works as white-box attack for their corresponding RPNs. Thus, the degradation for the respective object detector (highlighted in bold) is significantly larger. For example, the detection performance of {\em FR-v16} is degraded greatly from $59.2\%$ to $5.1\%$ at mAP $0.5$, and from $47.3\%$ to $3.1\%$ at mAP $0.7$. 

In comparison, the {\em RFCN} works as a black-box detector in our experiment. Observe that in the {\em RFCN} column of Table \ref{table:detection-perform}, the R-AP generated from {\em v16, mn, rn50, rn101, rn152} based RPN can effectively reduce the detection performance. In particular, the R-AP based on {\em rn101} can reduce the performance from $60.1\%$ to $52.0\%$ at mAP $0.5$, and from $50.0$ to $40.4\%$ at mAP $0.7$. 
The last row ($P$ for RFCN) shows the scaled accumulation of 5 perturbations ($p1, p2, p3, p4, p5$), which essentially represents the combination of effects learned from multiple networks that can notably degrade the performance of {\em RFCN} by $13.1\%$ at mAP 0.5 and $14.1\%$ at mAP 0.7.  

The performance of R-AP under different PSNR value on six object detectors are shown in Figure \ref{fig:det_plot_dB}. Observe that Gaussian noise (random) only produces small effects as the PSNR decreased. In contrast, the mAP curves of {\em v16, mn, rn50, rn101, rn152} in respective detector plots drop significantly compared to others. The black curve in each plot is the performance of accumulated perturbation $P$. We can see in pure black-box object detector {\em RFCN}, the accumulated perturbation curve drops notably and achieves the best results. 

We illustrate the visual performance of accumulated P on several object detectors in the first four rows of Figure \ref{fig:demos}. Due to the degradation of RPN after R-AP attack, the person in {\em FR-rn50} (b) is not detected. For the case of {\em FR-mn} (d), despite the target is correctly identified, the bounding box is disturbed to an undesired shape. 

\subsection{Instance Segmentation}
\label{subsection:inst_seg}

We evaluate the proposed R-AP on attacking two of the state-of-the-art instance segmentation methods in a black-box setting --- {\bf FCIS} \cite{FCIS} and Mask-RCNN ({\bf MR}) \cite{mask-rcnn}. 
Table \ref{table:instance-segment-perform} summarizes the performance degradation after applying R-AP at mAP 0.5 and 0.7. The $2^{nd}$ row of the table (random) shows a baseline by adding simple Gaussian noise as the perturbation, which is known to be ineffective in attacking, {\em i.e.} with only $<1\%$ performance decreased. In contrast, R-AP based on {\em v16, mn, rn50, rn101, rn152} each leads to large degradation of the performance. In particular, R-AP based on {\em rn101} degrades the performance of {\em FCIS} by $10.2\%$ at mAP $0.5$ and $9.8\%$ at mAP $0.7$, and reduces {\em MR} performance by $9.3\%$ at mAP $0.5$ and $7.8\%$ at mAP $0.7$. Notably, the accumulated perturbation $P$ degrades the performance of both methods --- a decrease of $15.1\%/13.2\%$ on {\em FCIS} and $16.7\%/13.8\%$ on {\em MR}. 

\begin{figure}[t]
        \centering
        \includegraphics[width=0.7\linewidth]{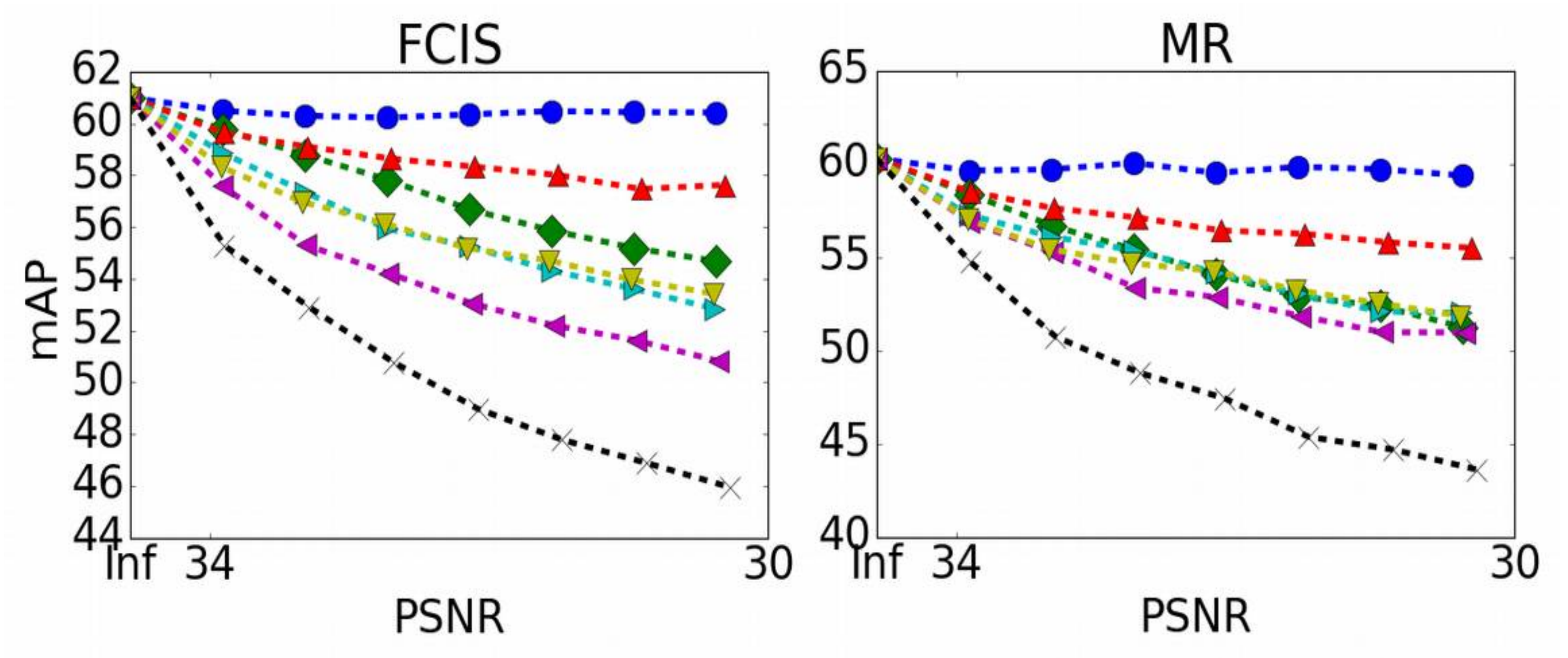}
        \captionof{figure}{\small \em Illustration of R-AP performance under different PSNR value at mAP 0.5 on 2 instance segmentation algorithms.}
        \label{fig:inst_plot_dB}
\end{figure}

Figure \ref{fig:inst_plot_dB} shows the performance evaluation of the R-AP black-box attack under different PSNR on the two instance segmentation methods. The blue curve corresponding to Gaussian noise is mostly flat, which shows the inefficacy of attack. In contrast, R-AP is effective for both instance segmentation methods. Notably, the black curve corresponding to the  accumulated perturbation $P$ achieves the largest degredation among all.

\begin{figure}[t]
	\centering
	\includegraphics[width=0.815\linewidth]{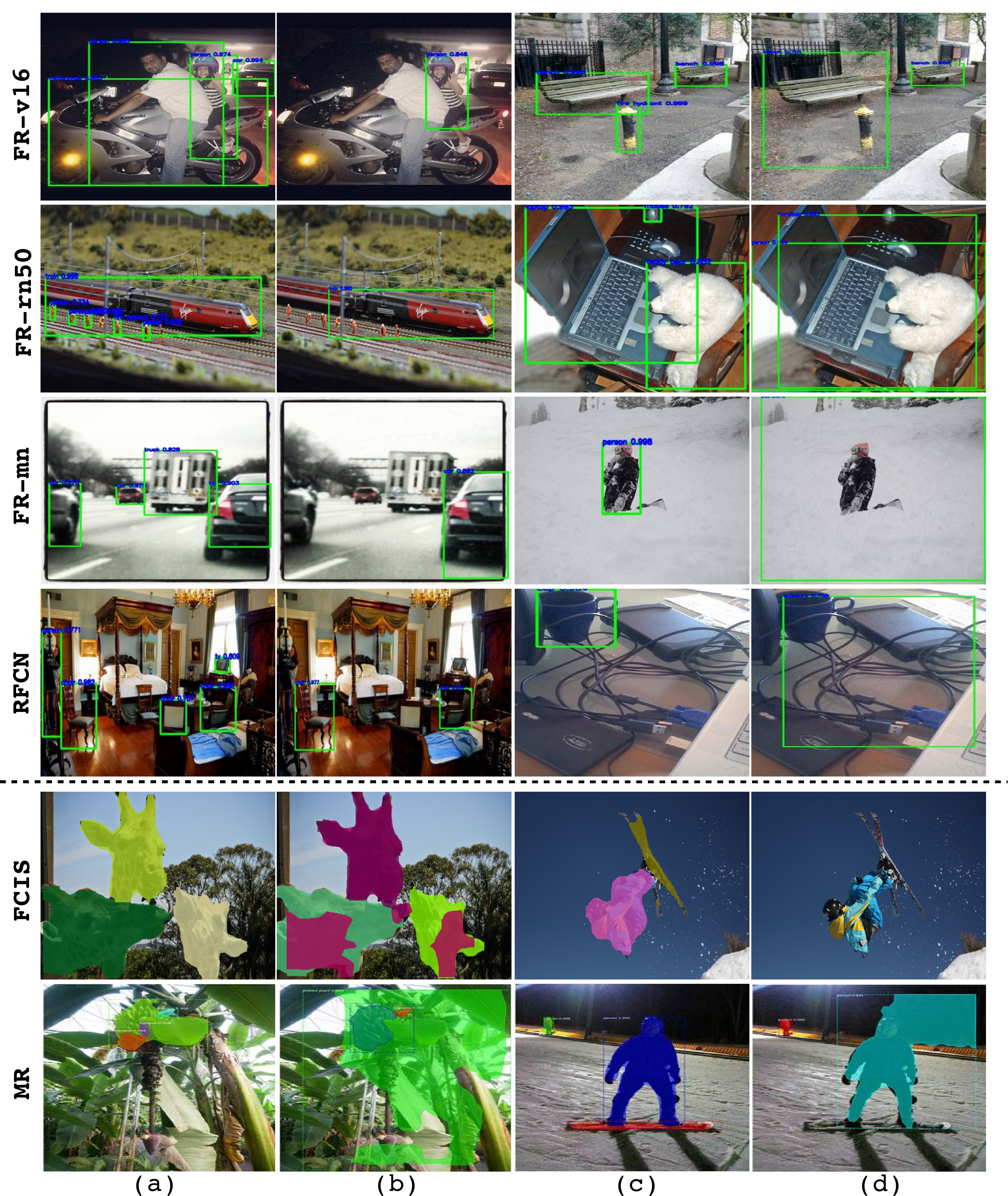}
	\caption{\small \em Visual results of the R-AP attack on several mainstream object detectors (first 4 rows) and instance segmentation algorithms (last 2 rows). 
		Columns (a,c) are the original results, and (b,d) are the R-AP attacked results.}
	\label{fig:demos}
\end{figure}

Visual illustration of the accumulated $P$ attack for instance segmentation is shown in the last two rows in Figure \ref{fig:demos}. Observe that the object instances are poorly segmented after the R-AP perturbation.  

\begin{table}[t]
      \small
      \centering
      \caption{\small \em Performance of black-box attack on 2 state-of-the-art instance segmentation algorithms at mAP $0.5$ and $0.7$. Lower value denotes better attacking performance.}
      \vspace{0.2cm}
      \begin{tabular}{| l | l| l |}
      \hline
          & {\bf FCIS} \cite{FCIS} & {\bf MR} \cite{mask-rcnn}\\
          \hline
          {\bf origin} & 61.0/45.8 & 60.3/45.6 \\
          \hline
          {\bf random} & 60.4/45.6 & 59.4/45.0 \\
          \hline
          {\bf v16} ($p_1$) & 54.7/40.0 & 51.3/38.1 \\
          \hline
          {\bf mn} ($p_2$)& 57.6/42.4 & 55.5/41.9 \\
          \hline
          {\bf rn50} ($p_3$)& 52.8/38.2 & 52.0/38.1 \\
          \hline
          {\bf rn101} ($p_4$)& 50.8/36.1 & 51.0/37.8 \\
          \hline
          {\bf rn152} ($p_5$)& 53.4/39.1 & 51.9/38.8 \\
          \hline
          ${\bf P} = \alpha \cdot \sum^5_{i=1} p_i $ & \bf 45.9/32.6 & \bf 43.6/31.8 \\
          \hline
      \end{tabular}
      \label{table:instance-segment-perform}
\end{table}

\section{Conclusion}

In this paper, we propose a robust adversarial perturbation (R-AP) method to attack deep proposal-based object detectors and instance segmentation algorithms. To the best of our knowledge, this work is the first to investigate the universal adversarial attack of the deep proposal-based models. Our method focuses on attacking Region Proposal Network (RPN), a component used in most deep proposal-based models, to universally affect the performance of their respective tasks. We describe a new loss function to not only disturb label prediction but also degrade shape regression. Evaluations on the MS COCO 2014 dataset shows that the R-AP can effectively attack several state-of-the-art object detectors and instance segmentation algorithms. 

Future work includes conducting further experiments on additional deep proposal-based models, including the part-based human pose detection. This work also opens up new opportunities on how to effectively improve the robustness of RPN-based networks.

\newpage
\small{
\bibliography{egbib}

\begin{thebibliography}{23}
\providecommand{\natexlab}[1]{#1}
\providecommand{\url}[1]{\texttt{#1}}
\expandafter\ifx\csname urlstyle\endcsname\relax
  \providecommand{\doi}[1]{doi: #1}\else
  \providecommand{\doi}{doi: \begingroup \urlstyle{rm}\Url}\fi

\bibitem[Chen and Gupta(2017)]{chen17implementation}
Xinlei Chen and Abhinav Gupta.
\newblock An implementation of {F}aster {RCNN} with study for region sampling.
\newblock \emph{arXiv 1702.02138}, 2017.

\bibitem[Dai et~al.(2016)Dai, Li, He, and Sun]{rfcn}
Jifeng Dai, Yi~Li, Kaiming He, and Jian Sun.
\newblock {R-FCN}: Object detection via region-based fully convolutional
  networks.
\newblock In \emph{NIPS}. 2016.

\bibitem[Everingham et~al.(2010)Everingham, Van~Gool, Williams, Winn, and
  Zisserman]{everingham2010pascal}
Mark Everingham, Luc Van~Gool, Christopher~KI Williams, John Winn, and Andrew
  Zisserman.
\newblock The {PASCAL} visual object classes ({VOC}) challenge.
\newblock \emph{IJCV}, 2010.

\bibitem[Evtimov et~al.(2018)Evtimov, Eykholt, Fernandes, Kohno, Li, Prakash,
  Rahmati, and Song]{evtimov2017robust}
Ivan Evtimov, Kevin Eykholt, Earlence Fernandes, Tadayoshi Kohno, Bo~Li, Atul
  Prakash, Amir Rahmati, and Dawn Song.
\newblock Robust physical-world attacks on deep learning models.
\newblock In \emph{CVPR}, 2018.

\bibitem[Girshick et~al.(2014)Girshick, Donahue, Darrell, and Malik]{rcnn}
Ross Girshick, Jeff Donahue, Trevor Darrell, and Jitendra Malik.
\newblock Rich feature hierarchies for accurate object detection and semantic
  segmentation.
\newblock In \emph{CVPR}, 2014.

\bibitem[Goodfellow et~al.(2015)Goodfellow, Shlens, and
  Szegedy]{goodfellow2014explaining}
Ian~J Goodfellow, Jonathon Shlens, and Christian Szegedy.
\newblock Explaining and harnessing adversarial examples.
\newblock In \emph{ICLR}, 2015.

\bibitem[He et~al.(2016)He, Zhang, Ren, and Sun]{he2016deep}
Kaiming He, Xiangyu Zhang, Shaoqing Ren, and Jian Sun.
\newblock Deep residual learning for image recognition.
\newblock In \emph{CVPR}, 2016.

\bibitem[He et~al.(2017)He, Gkioxari, Doll{\'a}r, and Girshick]{mask-rcnn}
Kaiming He, Georgia Gkioxari, Piotr Doll{\'a}r, and Ross Girshick.
\newblock Mask {R-CNN}.
\newblock In \emph{ICCV}, 2017.

\bibitem[Howard et~al.(2017)Howard, Zhu, Chen, Kalenichenko, Wang, Weyand,
  Andreetto, and Adam]{howard2017mobilenets}
Andrew~G Howard, Menglong Zhu, Bo~Chen, Dmitry Kalenichenko, Weijun Wang,
  Tobias Weyand, Marco Andreetto, and Hartwig Adam.
\newblock Mobilenets: Efficient convolutional neural networks for mobile vision
  applications.
\newblock \emph{arXiv 1704.04861}, 2017.

\bibitem[Kurakin et~al.(2017)Kurakin, Goodfellow, and
  Bengio]{kurakin2016adversarial}
Alexey Kurakin, Ian Goodfellow, and Samy Bengio.
\newblock Adversarial examples in the physical world.
\newblock In \emph{ICLR}, 2017.

\bibitem[Li et~al.(2017)Li, Qi, Dai, Ji, and Wei]{FCIS}
Yi~Li, Haozhi Qi, Jifeng Dai, Xiangyang Ji, and Yichen Wei.
\newblock Fully convolutional instance-aware semantic segmentation.
\newblock In \emph{CVPR}, 2017.

\bibitem[Lin et~al.(2014)Lin, Maire, Belongie, Hays, Perona, Ramanan,
  Doll{\'a}r, and Zitnick]{Lin2014MicrosoftCC}
Tsung-Yi Lin, Michael Maire, Serge~J. Belongie, James Hays, Pietro Perona, Deva
  Ramanan, Piotr Doll{\'a}r, and C.~Lawrence Zitnick.
\newblock Microsoft {COCO}: Common objects in context.
\newblock In \emph{ECCV}, 2014.

\bibitem[Lu et~al.(2017)Lu, Sibai, and Fabry]{lu2017adversarial}
Jiajun Lu, Hussein Sibai, and Evan Fabry.
\newblock Adversarial examples that fool detectors.
\newblock \emph{arXiv 1712.02494}, 2017.

\bibitem[Moosavi-Dezfooli et~al.(2016)Moosavi-Dezfooli, Fawzi, and
  Frossard]{moosavi2016deepfool}
Seyed-Mohsen Moosavi-Dezfooli, Alhussein Fawzi, and Pascal Frossard.
\newblock Deepfool: a simple and accurate method to fool deep neural networks.
\newblock In \emph{CVPR}, 2016.

\bibitem[Moosavi-Dezfooli et~al.(2017)Moosavi-Dezfooli, Fawzi, Fawzi, and
  Frossard]{moosavi2017universal}
Seyed-Mohsen Moosavi-Dezfooli, Alhussein Fawzi, Omar Fawzi, and Pascal
  Frossard.
\newblock Universal adversarial perturbations.
\newblock In \emph{CVPR}, 2017.

\bibitem[Papernot et~al.(2016)Papernot, McDaniel, Jha, Fredrikson, Celik, and
  Swami]{papernot2016limitations}
Nicolas Papernot, Patrick McDaniel, Somesh Jha, Matt Fredrikson, Z~Berkay
  Celik, and Ananthram Swami.
\newblock The limitations of deep learning in adversarial settings.
\newblock In \emph{EuroS\&P}, 2016.

\bibitem[Ren et~al.(2017)Ren, He, Girshick, and Sun]{faster-rcnn}
Shaoqing Ren, Kaiming He, Ross Girshick, and Jian Sun.
\newblock Faster {R-CNN}: Towards real-time object detection with region
  proposal networks.
\newblock \emph{TPAMI}, 2017.

\bibitem[Simonyan and Zisserman(2014)]{simonyan2014very}
Karen Simonyan and Andrew Zisserman.
\newblock Very deep convolutional networks for large-scale image recognition.
\newblock \emph{arXiv 1409.1556}, 2014.

\bibitem[Szegedy et~al.(2013)Szegedy, Zaremba, Sutskever, Bruna, Erhan,
  Goodfellow, and Fergus]{szegedy2013intriguing}
Christian Szegedy, Wojciech Zaremba, Ilya Sutskever, Joan Bruna, Dumitru Erhan,
  Ian Goodfellow, and Rob Fergus.
\newblock Intriguing properties of neural networks.
\newblock \emph{arXiv 1312.6199}, 2013.

\bibitem[Telagarapu et~al.(2011)Telagarapu, Naveen, Prasanthi, and
  Santhi]{telagarapu2011image}
Prabhakar Telagarapu, V~Jagan Naveen, A~Lakshmi Prasanthi, and G~Vijaya Santhi.
\newblock Image compression using {DCT} and wavelet transformations.
\newblock \emph{International Journal of Signal Processing, Image Processing
  and Pattern Recognition}, 2011.

\bibitem[Uijlings et~al.(2013)Uijlings, van~de Sande, Gevers, and
  Smeulders]{selective-search}
Jasper R.~R. Uijlings, Koen E.~A. van~de Sande, Theo Gevers, and Arnold W.~M.
  Smeulders.
\newblock Selective search for object recognition.
\newblock \emph{IJCV}, 2013.

\bibitem[Xie et~al.(2017)Xie, Wang, Zhang, Zhou, Xie, and
  Yuille]{xie2017adversarial}
Cihang Xie, Jianyu Wang, Zhishuai Zhang, Yuyin Zhou, Lingxi Xie, and Alan
  Yuille.
\newblock Adversarial examples for semantic segmentation and object detection.
\newblock In \emph{ICCV}, 2017.

\bibitem[Zeng et~al.(2017)Zeng, Liu, Qiu, Xie, Tai, Tang, and
  Yuille]{zeng2017adversarial}
Xiaohui Zeng, Chenxi Liu, Weichao Qiu, Lingxi Xie, Yu-Wing Tai, Chi~Keung Tang,
  and Alan~L Yuille.
\newblock Adversarial attacks beyond the image space.
\newblock \emph{arXiv 1711.07183}, 2017.

\end{thebibliography}
}

\end{document}